\documentclass[11pt]{article}

\usepackage[final]{acl}

\usepackage{times}
\usepackage{latexsym}

\usepackage[T1]{fontenc}

\usepackage[utf8]{inputenc}

\usepackage{microtype}

\usepackage{inconsolata}

\usepackage{graphicx}

\usepackage{
  url,
  comment,
  placeins,
  booktabs,
  multirow,
  array,
  tabularx,
  pgfplots,
  xcolor,
  pgfplotstable,
  caption,
  subcaption,
  amsmath,
  amssymb,
  amsfonts,
  algorithmic,
  textcomp,
  lineno,
  iftex,
  enumitem,
  pifont,
}

\ifXeTeX
\usepackage{xeCJK}
\else
\usepackage{CJKutf8}
\fi

\pgfplotsset{compat=1.18}
\definecolor{ytgreen}{HTML}{009E73}
\definecolor{hatorange}{HTML}{D55E00}
\definecolor{gridgray}{HTML}{D9D9D9}
\makeatletter
\renewcommand
\subsection{\@startsection{subsection}{2}{\z@}%
  {0.7ex \@plus .5ex \@minus .2ex}%
  {0.3ex \@plus .2ex}%
{\normalfont\normalsize\bfseries}}
\renewcommand\paragraph{\@startsection{paragraph}{4}{\z@}%
  {0.6ex \@plus .5ex \@minus .2ex}%
  {-1em}%
{\normalfont\normalsize\bfseries}} 
\makeatother

\title{Anchoring Speech with Semantics: A Multimodal Adapter Mechanism for Automatic Speech Recognition in Low-Resource Languages}

\author{
  \textbf{Kuan-Tang Huang\textsuperscript{1,3,*}},
  \textbf{Cheng-Yeh Yang\textsuperscript{1,*}},
  \textbf{Chien-Chun Wang\textsuperscript{1}}
  \\
  \textbf{Hung-Shin Lee\textsuperscript{1,$\dagger$}},
  \textbf{Hsin-Min Wang\textsuperscript{2}},
  \textbf{Berlin Chen\textsuperscript{1}}
  \\
  \\
  \textsuperscript{1}National Taiwan Normal University, Taiwan
  \\
  \textsuperscript{2}Academia Sinica, Taiwan
  \\
  \textsuperscript{3}EZAI, Taiwan
  \\
  \textsuperscript{*}Equal contribution
  \textsuperscript{$\dagger$}Corresponding author
}

\begin{document}
\maketitle

\begin{abstract}
  Low-resource ASR remains difficult because scarce transcripts provide limited supervised evidence for target-side generation. To address this gap, we propose SAMA-ASR, a lightweight adapter mechanism that augments the decoder with semantic anchors from auxiliary translations and an acoustic anchor from speech; in principle, the mechanism can be applied to similar encoder--decoder multitask speech models. Through cross-modal adaptation, SAMA-ASR conditions decoder states on translation-derived semantic embeddings and a speech embedding, combining utterance-level meaning with speech-grounded evidence before token prediction. At evaluation time, these semantic anchors can be generated automatically by an upstream speech-to-text translator rather than supplied as oracle translations. Experiments on two 30-hour datasets covering the low-resource Sinitic varieties Taiwanese Hokkien and Hakka show that SAMA-ASR improves over acoustic, prior prompt-based, and semantic-only translation-guided baselines and remains effective in practical automatic semantic-anchor settings; translator-capacity analyses show that useful semantic anchors can be produced by a compact ST model.
\end{abstract}

\section{Introduction}
Automatic Speech Recognition (ASR) aims to transcribe acoustic speech signals into accurate text sequences.
Although modern encoder--decoder architectures have substantially advanced ASR \cite{watanabe2017}, robust recognition for many low-resource languages and dialects remains a major gap because labeled speech is scarce \cite{chengExploringImpactData2025,yang2025,li2026}.

On the encoder side, multilingual speech pre-training can partially mitigate acoustic scarcity by transferring shared phonetic representations \cite{conneau2021,bapna2022}.
However, when only a few target-language transcripts are available, the decoder receives limited supervised evidence for target-variety lexical choices, orthographic conventions, common constructions, and meaning-compatible continuations.

To address this problem, we propose the Semantic-Aware Multimodal Adapter for ASR (SAMA-ASR)\footnote{Code is available at \url{https://github.com/610494/sama}.}, a lightweight decoder-side module that conditions ASR decoding on two complementary signals: semantic anchors encoded from auxiliary translations and an acoustic anchor derived from the speech signal.
Concretely, SAMA-ASR lets decoder hidden states query both translation-derived semantic embeddings and a speech embedding at each block, so local token decisions use utterance-level meaning while remaining tied to acoustic evidence.
This explicit acoustic path acts as a grounding constraint: it discourages semantically plausible but acoustically unsupported continuations when semantic anchors are noisy or incomplete.
Using translations as semantic anchors is practical because such dominant- or standard-language text appears across diverse low-resource speech settings, often when target-variety transcripts are unavailable \cite{xiao2023,pluss2023,zanonboito2022}.
From an implementation perspective, SAMA-ASR leaves the pretrained backbone frozen and, as an adapter, can in principle be attached to other end-to-end encoder--decoder ASR architectures.

This semantic-anchoring design is also useful for autoregressive decoding: semantic anchors are supplied before generation, providing global meaning context before partial hypotheses become reliable and reducing exposure to error accumulation \cite{bengio2015,arora2022}.

For practical inference without oracle translations, we introduce an upstream speech-to-text translation (ST) model to generate semantic anchors automatically.
Experiments show that a compact Whisper Small ST model already yields clear gains; compared with the frozen Whisper Medium ASR backbone used by SAMA-ASR (769M parameters), this 244M-parameter semantic-anchor generator keeps the added model cost moderate.
Closest to this setting, prior translation-guided ASR improves Taiwanese Hokkien ASR with Mandarin auxiliary translations, but assumes that oracle translations are provided and lacks an explicit acoustic-anchor path \cite{yang2026}.
In summary, the main contributions of this study are four-fold:
\begin{itemize}[leftmargin=*]
  \item \textbf{A semantic-anchoring formulation for low-resource ASR:} We recast extremely low-resource encoder--decoder ASR as a target-side generation problem and identify paired speech--translation data as a practical supervision source for improving decoder decisions when target-language transcripts are scarce.
  \item \textbf{A lightweight multimodal adapter architecture:} We introduce SAMA-ASR, a decoder-side adapter that lets autoregressive states attend to both translation-derived semantic anchors and an explicit acoustic anchor, improving semantic conditioning while preserving speech-grounded transcription and keeping the pretrained backbone frozen.
  \item \textbf{A realistic non-oracle inference pipeline:} We study automatic semantic-anchor generation with upstream ST models, showing that SAMA-ASR does not require oracle translations at test time and that compact translators can provide useful anchors with moderate added cost.
  \item \textbf{Cross-dataset validation and auxiliary-language analysis:} We validate SAMA-ASR on two low-resource Sinitic target varieties, Taiwanese Hokkien and Hakka, and provide analyses across data scales, translator capacities, auxiliary languages, and multilingual anchor compositions to clarify when semantic anchoring helps.
\end{itemize}

\section{Related Work}

\subsection{Low-Resource ASR}

Data scarcity remains a primary bottleneck for under-resourced ASR, including dialectal settings \cite{besacier2014,chengExploringImpactData2025,yang2025}.
Standard mitigation strategies include acoustic augmentation (e.g., speed perturbation and SpecAugment \cite{ko2015,park2019}) and resource construction from loosely aligned media or audiobook sources \cite{chen2020,yeroyan2024}.
Such resource-construction pipelines reduce annotation cost, but subtitle or long-form alignment can introduce label and segmentation noise.

A complementary direction designs ASR algorithms around the linguistic properties or resources of a specific target language, such as tone-aware modeling, character-set adaptation, or language-specific transfer strategies \cite{coto-solano2022,getman2024,peng2026}.
These language-specific studies show that target-language expertise can be valuable, but their assumptions may not transfer unchanged to languages with different phonology, scripts, orthographic conventions, or related-language resources.
In contrast, SAMA-ASR is not tied to a handcrafted linguistic feature or a single target language: our experiments show consistent gains on both Taiwanese Hokkien and Hakka, and our auxiliary-language analyses further suggest that semantic anchors can remain useful across translation languages when anchor quality and composition are favorable.

\paragraph{Joint ASR--ST and Parameter-Efficient Adaptation.}
Joint ASR--ST models exploit the complementarity between recognition and translation through jointly optimized decoders and cross-task information exchange \cite{leDualdecoderTransformerJoint2020}.
Separately, Meta-Adapter and LoRA-Whisper improve low-resource or multilingual ASR through parameter-efficient adaptation \cite{hou2021,songLoRAWhisperParameterEfficientExtensible2024}.
SAMA-ASR is complementary to these directions: rather than jointly optimizing ASR and ST objectives, it keeps the backbone frozen and injects translation-derived semantic and speech-derived acoustic evidence into the decoder, while its combination with LoRA shows that semantic--acoustic anchoring can further complement parameter-efficient adaptation.

\begin{figure*}[t]
  \centering
  \includegraphics[width=1.0\linewidth]{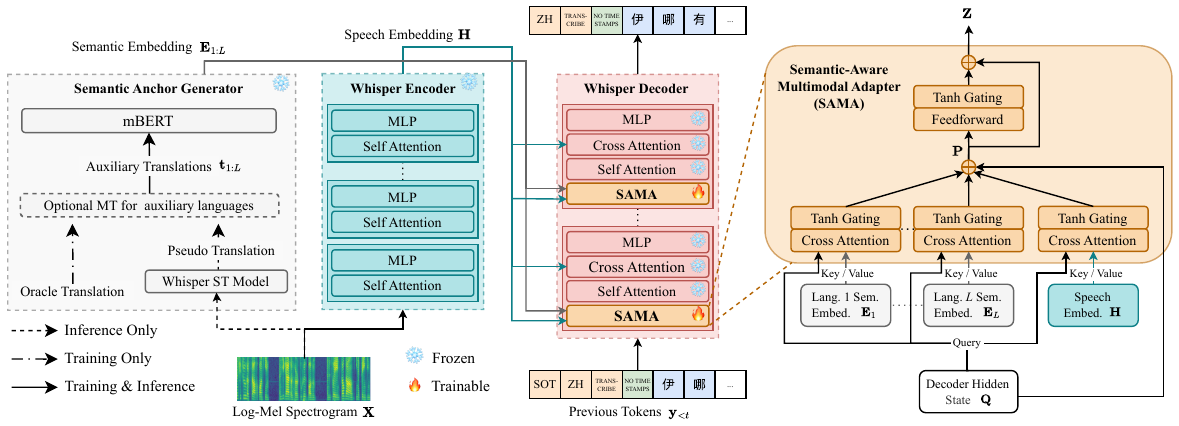}
  \vspace{-20pt}
  \caption{
    \textbf{Overview of the SAMA-ASR framework.}
    Snowflakes mark frozen modules and flames mark trainable modules.
    Automatic translations are produced by ST, encoded by mBERT, and fused with the frozen Whisper speech embedding through gated cross-attention.
    MT is used only in the optional multilingual diagnostic setting.
  }
  \vspace{-15pt}
  \label{fig:main}
\end{figure*}

\subsection{Text-Guided ASR}

Contextual biasing strengthens ASR decoders by injecting target-language priors or phrase-level constraints during recognition \cite{zhao2019,le2021,sun2023}.
These contextual-biasing methods are effective when the domain is known, but they usually require predefined phrase lists, target-language text, or task-specific biasing inventories, making them less suitable for open-domain low-resource ASR.
LLM-based ASR further brings reasoning and instruction-following ability \cite{chenIts2024,hsu2025}, but at billion-parameter-scale decoding or fusion cost.
Closest to our work, TG-ASR uses translation embeddings for Taiwanese Hokkien ASR \cite{yang2026}, but assumes translations are available at inference time.
SAMA-ASR instead targets a more general setting: the method does not require a predefined domain, phrase list, or oracle translation at test time, because semantic anchors can be generated automatically by a million-parameter-scale ST model.
SAMA-ASR then uses these anchors as utterance-level semantic guidance while adding an explicit acoustic-grounding path, so decoder decisions remain tied to the speech signal rather than relying on translation-only guidance.

\paragraph{Distillation vs. Anchoring.}
Text-guided speech distillation transfers linguistic knowledge from pretrained text or language-model teachers into ASR models during training \cite{choi2022,hentschel2024}.
Depending on the method, the teacher signal may be latent representation alignment or language-model token-probability targets, but it is used as a training-time supervision signal rather than an inference-time input.
By contrast, SAMA-ASR uses translated text as a runtime semantic anchor and combines it with an acoustic anchor through decoder-side adapter modules during decoding.

\paragraph{Multi-Source Text Fusion.}
SAMA-ASR also differs from source-text-assisted ASR for simultaneous interpretation, which uses an auxiliary source-language text encoder alongside the speech encoder \cite{taniguchi2022}.
That setting assumes source text is available at recognition time, whereas our practical setting must first generate auxiliary translations from speech.

\paragraph{Cross-Modal Adapter Design.}
SAMA-ASR's parameter-efficient design follows the same broad family as Flamingo-style gated cross-attention adapters and Whisper-Flamingo \cite{alayrac2022,rouditchenko2024}, while its multimodal grounding motivation is related to audio-visual ASR fusion work \cite{sterpu2018,ma2021}.
However, rather than fusing visual cues with speech, SAMA-ASR fuses semantic text anchors with an acoustic feature stream to separate meaning-level guidance from surface-form grounding.
Because SAMA-ASR attends over token-level semantic embeddings from mBERT \cite{devlin2019}, it preserves local translation structure instead of relying only on sentence-level semantic vectors such as LaBSE \cite{feng2022}.

\section{Methodology}

We present SAMA-ASR as a lightweight decoder-side adapter mechanism for enhancing low-resource ASR with semantic and acoustic anchors.
The central idea is: scarce target-language transcripts make the decoder's next-token distribution difficult to estimate reliably, so SAMA-ASR conditions decoder states on translation-derived semantic embeddings and a speech embedding, using the semantic embeddings for utterance-level meaning and the speech embedding to keep decoding grounded when translations are noisy or incomplete.
At inference time, semantic anchors are generated automatically by an upstream speech-to-text translation model, so deployment does not require oracle translations.
In our implementation, Whisper serves as the ASR backbone and mBERT serves as the semantic text encoder, although this adapter-based anchoring mechanism can in principle be attached to other encoder--decoder ASR architectures.
We first formalize this probabilistic framing, then describe semantic-anchor generation and the SAMA-ASR architecture with its training objective.

\subsection{Problem Framing and Objective}

Given a speech utterance $\mathbf{X}$ and a target transcription $\mathbf{y}=(y_1,\dots,y_N)$ of length $N$, a standard autoregressive ASR decoder models
\begin{equation}
  \label{eq:asr_factorization}
  p_{\boldsymbol{\theta}}(\mathbf{y}\mid\mathbf{X})
  =
  \prod_{i=1}^{N}p_{\boldsymbol{\theta}}(y_i\mid \mathbf{y}_{<i},\mathbf{X}).
\end{equation}
The ASR conditional distribution must learn both how acoustic evidence in $\mathbf{X}$ aligns with output tokens and an implicit target-language prior from the observed transcript history $\mathbf{y}_{<i}$.
In low-resource settings, the paired transcript set is too small to estimate this distribution reliably:
\begin{equation}
  \label{eq:low_resource_mismatch}
  \widehat{p}_{\mathrm{low}}(y_i\mid \mathbf{y}_{<i},\mathbf{X})
  \not\approx
  p^{*}(y_i\mid \mathbf{y}_{<i},\mathbf{X}).
\end{equation}
Here $p^{*}$ denotes the true target conditional distribution that the model is intended to approximate, while $\widehat{p}_{\mathrm{low}}$ denotes the distribution learned from limited transcripts.
This distributional mismatch reflects a limited-sample estimation error: scarce transcripts provide an incomplete sample of the underlying target-language distribution, so the decoder may place excessive probability mass on tokens that are locally plausible in the observed data but inconsistent with the utterance meaning or the speech signal.

To reduce this mismatch while preserving the ASR target, SAMA-ASR augments the decoder's conditioning information.
For an utterance, we define a set of anchors
\begin{equation}
  \label{eq:anchor_set}
  \mathcal{A}(\mathbf{X},\mathcal{T})
  =
  \{\mathbf{E}_1,\dots,\mathbf{E}_L,\mathbf{H}\},
\end{equation}
where $\mathcal{T}=\{\mathbf{t}_1,\dots,\mathbf{t}_L\}$, $L$ is the number of auxiliary translations, $\mathbf{E}_l$ is the semantic embedding of an auxiliary translation $\mathbf{t}_l$, and $\mathbf{H}$ is the speech embedding produced by the frozen speech encoder.
For brevity, we write $\mathcal{A}$ for $\mathcal{A}(\mathbf{X},\mathcal{T})$ when the utterance and translations are clear from context.
In the main setting, $L=1$ with Mandarin as the auxiliary translation, but the formulation can be extended to $L>1$ with multiple auxiliary translations; we evaluate this extension in Appendix \ref{app:typological_and_multilingual}.
The adapted decoder therefore estimates
\begin{equation}
  \label{eq:sama_decoder}
  \begin{aligned}
    p_{\boldsymbol{\theta},\boldsymbol{\psi}}(\mathbf{y}\mid\mathbf{X},\mathcal{A})
    &=
    \prod_{i=1}^{N}
    p_{\boldsymbol{\theta},\boldsymbol{\psi}}(y_i\mid \mathbf{y}_{<i},\mathbf{X},\mathcal{A}),
  \end{aligned}
\end{equation}
where $\boldsymbol{\theta}$ denotes the frozen ASR backbone parameters and $\boldsymbol{\psi}$ denotes the trainable SAMA-ASR parameters.
Semantic anchors narrow the distribution toward meaning-compatible hypotheses, while the acoustic anchor prevents the model from drifting toward translation-like or acoustically unsupported outputs.
Together, these anchors compensate for the limited-sample distributional bias by adding semantic evidence while preserving acoustic grounding for the same ASR target.
Because this anchor evidence is available throughout autoregressive decoding, it also mitigates early-decoding fragility after generation begins: the first few continuation tokens need not rely only on a very short and potentially unreliable history $\mathbf{y}_{<i}$.
After early mistakes, the anchors continue to provide utterance-level semantic and acoustic references, reducing the decoder's dependence on its own erroneous partial history.

\subsection{Semantic Anchor Generation}
\label{ssec:semantic_anchor_generation}

The left side of Figure \ref{fig:main} shows how SAMA-ASR obtains semantic anchors before ASR decoding.
The auxiliary translation $\mathbf{t}_1$ is associated with the input speech; in our Taiwanese Hokkien and Hakka experiments, $\mathbf{t}_1$ is instantiated as Mandarin because paired Mandarin translations are available.
Accordingly, SAMA-ASR assumes paired speech--translation data during adaptation and does not directly apply when auxiliary translations are entirely unavailable.
During SAMA-ASR training, $\mathbf{t}_1$ is the paired oracle translation from the training split, which provides clean semantic conditioning rather than a pseudo-transcription target.
Using oracle translations during adaptation avoids propagating translator errors into the adapter, analogous to teacher forcing with clean conditioning context.
Appendix \ref{app:training_anchor_source} supports this choice in our ablation setting: the train--test translation-source mismatch is small with 30 hours of paired data, while oracle-translation training becomes more robust than pseudo-translation training when paired data is reduced.
For practical inference without human-provided translations, we start from a pretrained Whisper model and fine-tune it independently for speech-to-text translation, using the same training speech utterances as ASR adaptation but the corresponding auxiliary translations as output labels.
We optimize this ST generator with a standard cross-entropy translation objective and run it on each test utterance to produce a pseudo translation.

For the $L>1$ diagnostic analyses, the Mandarin auxiliary translations, instantiated by oracle translations in those analyses, are translated into English, Hindi, Spanish, and French with SeamlessM4T \cite{barrault2023}, forming multilingual auxiliary translations $\mathcal{T}=\{\mathbf{t}_1,\dots,\mathbf{t}_L\}$ with up to five total translations.
A frozen mBERT encoder then converts each auxiliary translation into token-level semantic embeddings $\mathbf{E}_{1:L}$ \cite{devlin2019}.
Oracle-translation settings use the same mBERT encoding step but replace the ST-generated pseudo translation with the oracle Mandarin translation.

\subsection{SAMA-ASR Architecture}
\label{ssec:sama_architecture}

Given semantic embeddings $\mathbf{E}_{1:L}$, the speech input $\mathbf{X}$, and the anchor set $\mathcal{A}=\{\mathbf{E}_1,\dots,\mathbf{E}_L,\mathbf{H}\}$ defined above, SAMA-ASR keeps the original Whisper encoder and decoder frozen and inserts an additional SAMA-ASR module into each decoder block, as shown in Figure \ref{fig:main}.
The Whisper encoder maps $\mathbf{X}$ to a speech embedding $\mathbf{H}$, while the semantic-anchor generator described in Section \ref{ssec:semantic_anchor_generation} provides the semantic embeddings.

Within each decoder block, SAMA-ASR is inserted before decoder self-attention, so the current decoder state can query both translation-derived semantic embeddings and the speech embedding before the frozen decoder updates its autoregressive token-history representation.
Let $\mathbf{Q}^{(b)}\in\mathbb{R}^{S\times D}$ denote the decoder hidden states at the SAMA-ASR insertion point of block $b$, where $S$ is the output-side sequence length and $D$ is the decoder width.
Before attention, each anchor representation $\mathbf{A}_k\in\mathcal{A}$ is mapped to the decoder width.
SAMA-ASR then uses parallel cross-attention branches so that each anchor can be queried and gated independently.
For an anchor $\mathbf{A}_k$, the decoder state obtains
\begin{equation}
  \label{eq:sama_cross_attention}
  \mathbf{C}_k
  =
  \operatorname{CrossAttn}_k(\mathbf{Q}^{(b)},\mathbf{A}_k,\mathbf{A}_k),
\end{equation}
where $\mathbf{A}_k$ provides both keys and values, and each branch has its own trainable attention projections.
The contribution of each branch is controlled by a learnable scalar gate:
\begin{equation}
  \label{eq:sama_gate}
  \widehat{\mathbf{C}}_k
  =
  \tanh(\alpha_k)\mathbf{C}_k.
\end{equation}
The gated contexts are summed with the original decoder state to form the intermediate state
\begin{equation}
  \label{eq:sama_aggregation}
  \mathbf{P}^{(b)}
  =
  \mathbf{Q}^{(b)}+
  \sum_{\mathbf{A}_k\in\mathcal{A}}\widehat{\mathbf{C}}_k.
\end{equation}
Finally, SAMA-ASR applies a gated feed-forward adapter branch,
\begin{equation}
  \label{eq:sama_ffn}
  \mathbf{Z}^{(b)}
  =
  \mathbf{P}^{(b)}+\tanh(\alpha_{\mathrm{ffn}})\operatorname{FFN}(\mathbf{P}^{(b)}),
\end{equation}
and passes $\mathbf{Z}^{(b)}$ to the frozen self-attention sublayer of the original Whisper decoder block.
Following Flamingo-style gated cross-attention/dense blocks \cite{alayrac2022} and their Whisper adaptation \cite{rouditchenko2024}, all gate parameters $\{\alpha_k,\alpha_{\mathrm{ffn}}\}$ are initialized to zero, making each inserted SAMA-ASR module initially behave as an identity adapter, $\mathbf{Z}^{(b)}\approx\mathbf{Q}^{(b)}$, before it learns to use semantic and acoustic evidence.
The SAMA-ASR adapter design augments rather than replaces Whisper's original frozen encoder--decoder attention: the pretrained acoustic pathway remains intact, while the additional acoustic branch provides a trainable gated memory that can counterbalance noisy semantic anchors before autoregressive self-attention.

\paragraph{Training Objective.}
\label{ssec:training_objective}
Following parameter-efficient low-resource ASR adaptation work \cite{hou2021}, we optimize only the SAMA-ASR modules and their projection layers to reduce overfitting under scarce target-variety transcripts.
SAMA-ASR is trained with a standard cross-entropy loss on the target-language transcription:
\begin{equation}
  \label{eq:asr_loss}
  \mathcal{L}_{\mathrm{ASR}}
  =
  -\sum_{i=1}^{N}
  \log p_{\boldsymbol{\theta},\boldsymbol{\psi}}(y_i\mid \mathbf{y}_{<i},\mathbf{X},\mathcal{A}).
\end{equation}
\section{Experimental Setup}
\subsection{Datasets and Metrics}

We evaluate SAMA-ASR on two low-resource Sinitic ASR settings: Taiwanese Hokkien and Hakka.
Both settings provide paired dialect speech and Mandarin auxiliary translations, allowing us to study the realistic question of how much paired data is available rather than assuming a separate large resource for the translation component.
For both datasets, transcriptions follow the corpus convention of Traditional Chinese character-based notation.
We report character error rate (CER) over the corpus-provided character sequence, since Taiwanese Hokkien and Hakka lack standardized word segmentation and explicit word delimiters.

\paragraph{Taiwanese Hokkien (YT-THDC).}
YT-THDC is a previously introduced Taiwanese Hokkien corpus consisting of YouTube drama audio paired with open-caption Mandarin translations \cite{yang2026}.
These Mandarin translations are used as loosely aligned auxiliary translations rather than deterministic pseudo-transcripts \cite{chen2020}.

\paragraph{Taiwanese Hakka (HAT).}
For Hakka, we construct a controlled 30-hour low-resource subset from the Hakka Across Taiwan corpus \cite{liao2023}, matching the scale of YT-THDC while preserving dialectal variation.
Appendix \ref{app:experimental_details} provides dataset construction details and the split statistics in Table \ref{tab:dataset_statistic} for the controlled 30-hour low-resource setting used for both datasets.

\begin{table*}[t]
  \centering
  \small
  \setlength{\tabcolsep}{1.5pt}
  \resizebox{0.98\textwidth}{!}{%
    \begin{tabular}{@{}llll ccc ccc@{}}
      \toprule
      \multirow{2}{*}{\textbf{Method}} & \multirow{2}{*}{\textbf{Train}} & \multirow{2}{*}{\textbf{Eval}} & \multirow{2}{*}{\textbf{Aud.}} & \multicolumn{3}{c}{\textbf{YT-THDC}} & \multicolumn{3}{c}{\textbf{HAT}} \\
      \cmidrule(lr){5-7} \cmidrule(lr){8-10}
      & & & & \textbf{CER} $\downarrow$ & \textbf{Rel. Impr.} $\uparrow$ & \textbf{Rel. Impr.} $\uparrow$ & \textbf{CER} $\downarrow$ & \textbf{Rel. Impr.} $\uparrow$ & \textbf{Rel. Impr.} $\uparrow$ \\
      & & & & & \textbf{vs. Base} & \textbf{vs. Audio} & & \textbf{vs. Base} & \textbf{vs. Audio} \\
      \midrule
      \multicolumn{10}{l}{\textit{\textbf{Baseline}}} \\
      Self Attn. & -- & -- & No & 33.49 & -- & -- & 27.78 & -- & -- \\
      Cross Attn. & -- & -- & Yes & 32.89 & 1.79 & -- & 26.58 & 4.32 & -- \\
      \midrule
      \multicolumn{10}{l}{\textit{\textbf{Prior Text-Guided Baselines + SAMA-ASR} (Practical Evaluation)}} \\
      Self Attn. + Prompt \cite{peng2023} & -- & Auto & No & 27.24 & 18.66 & 17.18 & 25.45 & 8.39 & 4.25 \\
      TG-ASR \cite{yang2026} & Oracle & Auto & No & 24.38 & 27.20 & 25.87 & 24.97 & 10.12 & 6.06 \\
      SAMA-ASR & Oracle & Auto & Yes & \textbf{23.48} & \textbf{29.89} & \textbf{28.61} & \textbf{22.48} & \textbf{19.08} & \textbf{15.43} \\
      \midrule
      \multicolumn{10}{l}{\textit{\textbf{LoRA Baseline + SAMA-ASR}}} \\
      LoRA & -- & -- & -- & 23.60 & 29.53 & 28.25 & 19.30 & 30.53 & 27.39 \\
      LoRA + SAMA-ASR & Oracle & Auto & Yes & \textbf{20.64} & \textbf{38.37} & \textbf{37.25} & \textbf{16.82} & \textbf{39.45} & \textbf{36.72} \\
      \midrule
      \multicolumn{10}{l}{\textit{\textbf{Diagnostic Oracle Evaluation}}} \\
      TG-ASR & Oracle & Oracle & No & 20.15 & 39.83 & 38.74 & 24.28 & 12.60 & 8.65 \\
      SAMA-ASR & Oracle & Oracle & Yes & \textbf{16.79} & \textbf{49.87} & \textbf{48.95} & \textbf{19.63} & \textbf{29.34} & \textbf{26.15} \\
      \bottomrule
  \end{tabular}}
  \vspace{-5pt}
  \caption{
    \textbf{Main ASR results on YT-THDC and HAT.}
    CER values are percentages; lower is better.
    Relative-improvement columns report CER reduction rates (\%) against the context-only Self Attn. baseline (``vs. Base'') and the audio-only Cross Attn. baseline (``vs. Audio''); higher is better.
    ``Oracle''/``Auto'' denote the auxiliary-translation source, and ``Aud.'' indicates whether an explicit acoustic anchor is used.
    Bold practical SAMA-ASR results are significant at $p < 0.05$ by \citet{bisani2004}: standalone SAMA-ASR is compared with all non-LoRA baselines above it, and LoRA+SAMA-ASR is compared with LoRA.
  }
  \label{tab:main_results}
  \vspace{-15pt}
\end{table*}

\subsection{Baselines and Protocol}
\label{ssec:baselines}

Table \ref{tab:main_results} contains three comparisons: audio-only adaptation, prior text-guided baselines under the same practical automatic-translation setting, and composition with LoRA.
Self Attn.+Prompt supplies automatic Mandarin through Whisper's native prompt interface \cite{peng2023}, TG-ASR serves as a strong semantic-only baseline by encoding the same text as decoder-side anchors \cite{yang2026}, and SAMA-ASR further adds the acoustic anchor.
The prompt baseline uses the same automatic Mandarin translation source as the practical TG-ASR and SAMA-ASR rows; implementation details are in Appendix \ref{app:experimental_details}.
For the practical comparisons, the upstream ST generator only produces auxiliary Mandarin strings shared by Self Attn.+Prompt, TG-ASR, and SAMA-ASR; no oracle test translations are used, and the generator parameters do not participate in ASR decoding.

\section{Results and Discussion}

\subsection{Main Results}

Following Sec. \ref{ssec:baselines}, Table \ref{tab:main_results} tests whether semantic anchors help beyond audio-only and prior prompt-based baselines, whether acoustic grounding improves over the strong semantic-only TG-ASR baseline, and whether SAMA-ASR composes with LoRA.
All rows in the table use per-utterance greedy decoding with beam size 1 and temperature 0.
Because practical ASR inference must generate hypotheses without reference-token history, we evaluate all systems in the same free-running regime.
This free-running evaluation includes TG-ASR, whose original report used teacher-forced evaluation; the TG-ASR numbers here are therefore intended for controlled comparison with SAMA-ASR rather than direct comparison with the originally reported teacher-forced scores.
\paragraph{Unimodal vs. Multimodal.}
Standard acoustic adaptation yields only limited gains ($\le$4.32\% relative improvement), suggesting that sparse acoustic supervision alone does not sufficiently reshape decoder-side prediction.
The prior prompt-based baseline shows that automatic Mandarin prompts help, while the strong TG-ASR baseline reduces CERs further to 24.38\% on YT-THDC and 24.97\% on HAT, indicating that adapter-level semantic anchoring is stronger than prompt-based conditioning.
Adding the acoustic anchor in SAMA-ASR further improves the practical setting to 23.48\% and 22.48\%, indicating that acoustic grounding supplies complementary fine-grained evidence rather than merely correcting noisy translations. Appendix \ref{sec:appendix_case_study} provides a qualitative example.

\paragraph{Oracle Trends.}
Oracle evaluation clarifies the role of anchor fidelity.
On YT-THDC, replacing automatic Mandarin translations with oracle Mandarin translations improves TG-ASR from 24.38\% to 20.15\%, and adding the acoustic anchor in SAMA-ASR further reduces CER to 16.79\%.
The larger oracle-condition gain suggests that reliable semantic anchors amplify acoustic grounding: once the semantic stream narrows the hypothesis space, acoustic cues can more effectively resolve remaining surface-form ambiguities.
On HAT, oracle TG-ASR (24.28\%) still trails practical SAMA-ASR (22.48\%), but oracle SAMA-ASR reaches 19.63\%, again showing that acoustic grounding remains essential when translation semantics diverge from colloquial dialectal forms.

\paragraph{Compatibility with a Strong PEFT Baseline.}
LoRA is a strong PEFT baseline, reaching 23.60\% CER on YT-THDC and 19.30\% on HAT, so standalone SAMA-ASR is not meant to replace it.
Instead, SAMA-ASR acts as a compositional semantic-acoustic module: LoRA+SAMA-ASR further reduces CER to 20.64\% and 16.82\%, corresponding to 12.54\% and 12.85\% relative improvements over LoRA alone.

{
  \begin{table}[t]
    \small
    \centering
    \begin{tabularx}{0.98\columnwidth}{@{}Xcc@{}}
      \toprule
      \textbf{Acoustic-Anchor Condition} & \textbf{YT-THDC} & \textbf{HAT} \\
      \midrule
      Self Attn. baseline & 33.49 & 27.78 \\
      Cross Attn. baseline & 32.89 & 26.58 \\
      Random Noise (matched mean/std) & 42.62 & 30.91 \\
      Shuffle Across Utterances & 45.49 & 35.59 \\
      \textbf{SAMA-ASR} & \textbf{23.48} & \textbf{22.48} \\
      \bottomrule
    \end{tabularx}
    \caption{
      \textbf{Controlled acoustic-anchor ablation.}
      CERs (\%; lower is better).
      The controlled variants retain the complete SAMA-ASR architecture and trainable parameter count; the assigned corruption is used during both training and evaluation.
    }
    \label{tab:controlled_acoustic_anchor_ablation}
    \vspace{-10pt}
  \end{table}

  \subsection{Controlled Acoustic-Anchor Ablation}
  \label{ssec:controlled_acoustic_anchor_ablation}

  To verify that the acoustic anchor contributes speech-grounded information rather than merely additional model capacity, we keep the complete SAMA-ASR architecture and trainable parameter count fixed while corrupting only the acoustic-anchor input.
  \textbf{Random Noise} replaces the acoustic anchor with Gaussian noise matched to the original features' mean and standard deviation, whereas \textbf{Shuffle Across Utterances} preserves real acoustic features but assigns them to different utterances within each mini-batch.
  Each corruption is applied during both training and evaluation.

  Table \ref{tab:controlled_acoustic_anchor_ablation} shows that both controls substantially degrade performance relative to SAMA-ASR.
  Random noise raises CER to 42.62\% on YT-THDC and 30.91\% on HAT, while utterance shuffling further degrades performance to 45.49\% and 35.59\%, compared with 23.48\% and 22.48\% for SAMA-ASR.
  Both controls also fall below the Self Attn. and Cross Attn. baselines, showing that the acoustic branch alone does not explain the gain.
  The shuffled condition is particularly diagnostic because it preserves real acoustic representations and their statistics while breaking only utterance alignment.
  Together, these results isolate the source of SAMA-ASR's gain: correctly aligned speech-derived evidence rather than generic acoustic features or additional capacity.
}

\subsection{Autoregressive Cold-Start Analysis}

\begin{figure}[t]
  \centering
  \includegraphics[width=\linewidth]{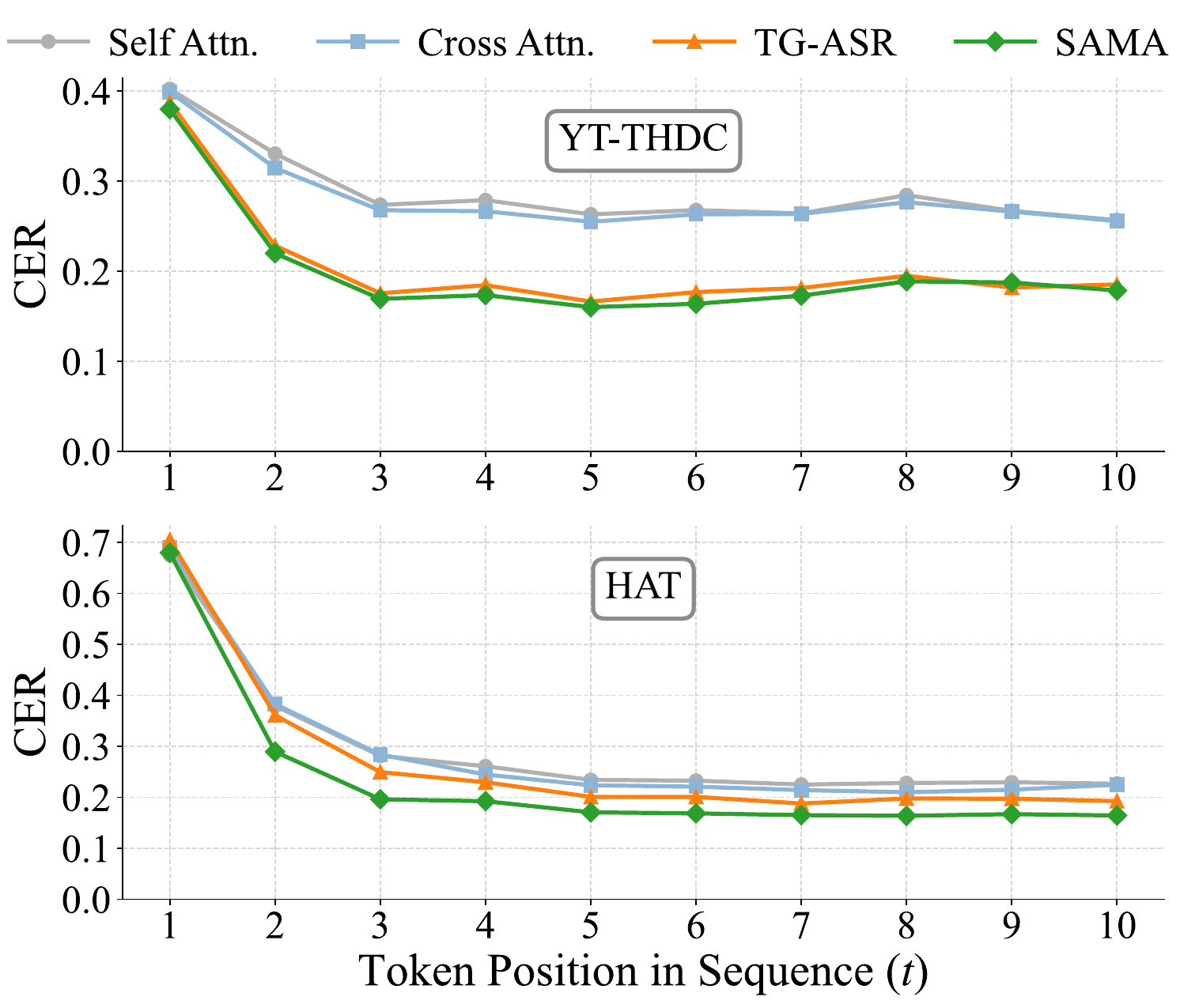}
  \vspace{-20pt}
  \caption{
    \textbf{Position-wise CER analysis of autoregressive cold-start fragility.}
    Average CER is reported over the first ten target-token positions on the test sets.
    TG-ASR is the strong semantic-only translation-guided baseline, while SAMA-ASR adds the acoustic-anchor path.
  }
  \label{fig:cold_start_er}
  \vspace{-10pt}
\end{figure}

Figure \ref{fig:cold_start_er} examines the autoregressive vulnerability motivating SAMA-ASR by measuring CER by output-token position against non-semantic self-attention, audio-only cross-attention, and the strong semantic-only TG-ASR baseline.
Both datasets show a clear cold-start pattern: the first few positions are much harder than later ones, confirming that early decisions are fragile when little target-side history is available.
SAMA-ASR does not substantially improve the first position, where the initial surface token is still strongly tied to acoustic onset and utterance alignment.
Its advantage emerges from positions 2--3 onward, supporting our framing that semantic anchors help once the decoder begins conditioning on a fragile partial history rather than acting as an oracle for the first token.
Compared with TG-ASR, SAMA-ASR shows the clearest additional benefit on HAT, where the acoustic-anchor path yields a stronger early-position separation than semantic-only conditioning.
On YT-THDC, the two translation-guided methods are closer, indicating a smaller but still favorable incremental effect from acoustic grounding rather than position-wise dominance.

\subsection{Robustness}

\begin{figure}[t]
  \centering
  \includegraphics[width=1.0\linewidth]{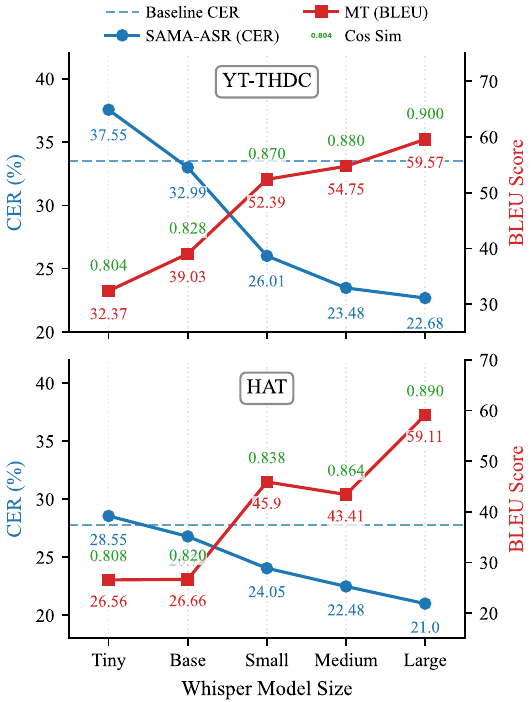}
  \vspace{-20pt}
  \caption{
    \textbf{Evaluation-time translator-capacity analysis.}
    Values are SAMA-ASR CERs using automatic Mandarin translations from translators of different sizes.
    Dashed lines denote context-only CER baselines; red values report translator BLEU, and green values report average cosine similarity between mBERT embeddings of automatic and oracle Mandarin translations.
  }
  \vspace{-15pt}
  \label{fig:scalability_analysis}
\end{figure}

We evaluate robustness under two practical constraints: varying the evaluation-time translator capacity while fixing ASR adaptation, and reducing the paired-data budget used for both ASR adaptation and translator training.
The latter setting is stricter because data scarcity weakens the acoustic adapter, the upstream translator, and the resulting auxiliary translations at the same time.

\paragraph{Upstream Translation Robustness.}
Figure \ref{fig:scalability_analysis} shows that SAMA-ASR does not require an oracle-quality translator at inference time.
Base and larger translators outperform the context-only baseline on both datasets; even with modest BLEU, mBERT-space similarity to oracle translations remains high.
BLEU is not a monotonic proxy for downstream utility: on HAT, the Small translator has higher BLEU than Medium (45.90 vs. 43.41), yet its lower mBERT cosine similarity (0.838 vs. 0.864) coincides with worse CER (24.05 vs. 22.48).
The BLEU--CER mismatch suggests that SAMA-ASR is more sensitive to semantic-neighborhood compatibility in the mBERT space consumed by the adapter than to surface n-gram overlap alone. Appendix \ref{app:semantic_neighborhood_visualization} visualizes these mBERT neighborhoods.

\paragraph{End-to-End Data-Scarcity Robustness.}
Table \ref{tab:data_scale_comparison} evaluates a stricter scenario where the same paired-data subset trains both the ASR adapter and the translator.
With 1 hour and 10 hours, SAMA-ASR gives the best CERs on both datasets.
The data-scarcity results show that semantic anchors need not be oracle translations, but they must remain sufficiently meaning-compatible for the decoder-side benefit to outweigh translation noise.

\begin{table}[t]
  \small
  \centering
  \begin{tabularx}{0.98\columnwidth}{@{}XXcc@{}}
    \toprule
    \textbf{Method} & \textbf{Data} & \textbf{YT-THDC} & \textbf{HAT} \\
    \midrule
    Self Attention & 10 min & \textbf{61.32} & 96.44 \\
    Cross Attention & 10 min & 66.18 & 78.27 \\
    SAMA-ASR & 10 min & 65.63 & \textbf{72.67} \\
    \midrule
    Self Attention & 1 hr & 55.92 & 68.37 \\
    Cross Attention & 1 hr & 54.97 & 57.20 \\
    SAMA-ASR & 1 hr & \textbf{53.45} & \textbf{54.68} \\
    \midrule
    Self Attention & 10 hr & 38.10 & 35.99 \\
    Cross Attention & 10 hr & 39.04 & 33.61 \\
    SAMA-ASR & 10 hr & \textbf{30.04} & \textbf{30.93} \\
    \bottomrule
  \end{tabularx}
  \caption{
    \textbf{Paired-data budget analysis.}
    CERs (\%) when the same subset trains both ASR adaptation and the translator; SAMA-ASR uses automatic Mandarin translations at evaluation.
  }
  \label{tab:data_scale_comparison}
  \vspace{-10pt}
\end{table}

{
  \paragraph{Practical Boundary Conditions.}
  The benefits of semantic anchoring are not uniform under the most resource-constrained conditions.
  With a Tiny upstream translator, SAMA-ASR can fall below the context-only baseline, indicating that anchors with insufficient semantic fidelity may not provide reliable guidance.
  Similarly, in the 10-minute paired-data setting, SAMA-ASR performs best on HAT and improves over Cross Attn. on YT-THDC, but trails Self Attn. on YT-THDC (65.63\% vs. 61.32\% CER).
  These cases delineate the practical operating boundary of SAMA-ASR: the method requires either sufficient paired supervision or an upstream translator whose outputs remain meaning-compatible enough to offset uncertainty in low-resource decoding.
}

\subsection{Semantic Prior Ablation}

Table \ref{tab:text_encoder_comparison} isolates the role of the text encoder in the audio-enhanced SAMA-ASR setting.
Replacing frozen mBERT with random embeddings substantially degrades performance, and Whisper embeddings perform even worse, showing that the gains do not come merely from adding another conditioning stream.
Rather, SAMA-ASR depends on cross-lingual semantic structure in the mBERT semantic encoder, which constrains decoder hypotheses while acoustic evidence determines the final surface form.

\subsection{Auxiliary Languages}

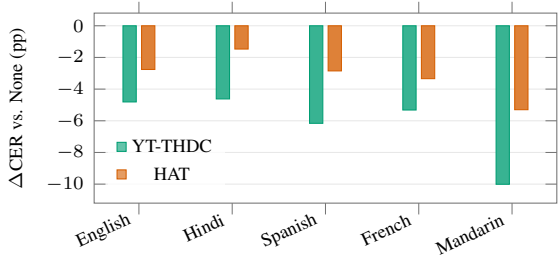
\begin{figure}[t]
  \centering
  \begin{tikzpicture}
    \begin{axis}[
        width=\columnwidth,
        height=4.1cm,
        ybar,
        bar width=5pt,
        ymin=-11.2,
        ymax=0.8,
        ymajorgrids,
        grid style={gridgray!60},
        ylabel={$\Delta$CER vs. None (pp)},
        symbolic x coords={English,Hindi,Spanish,French,Mandarin},
        xtick=data,
        xticklabel style={rotate=25,anchor=east,font=\scriptsize},
        ytick={-10,-8,-6,-4,-2,0},
        yticklabel style={font=\scriptsize},
        ylabel style={font=\scriptsize},
        legend style={
          at={(0.03,0.04)},
          anchor=south west,
          legend columns=1,
          draw=none,
          fill=white,
          fill opacity=0.78,
          text opacity=1,
          font=\scriptsize,
          /tikz/every even column/.append style={column sep=2pt},
        },
        legend image code/.code={\draw[#1,fill=#1] (0cm,-0.04cm) rectangle (0.12cm,0.08cm);},
        enlarge x limits=0.12,
        axis line style={black!55},
        tick style={black!55},
      ]
      \addplot+[fill=ytgreen!78,draw=ytgreen] coordinates {
        (English,-4.81) (Hindi,-4.62) (Spanish,-6.16) (French,-5.32) (Mandarin,-10.01)
      };
      \addplot+[fill=hatorange!78,draw=hatorange] coordinates {
        (English,-2.76) (Hindi,-1.47) (Spanish,-2.85) (French,-3.34) (Mandarin,-5.30)
      };
      \legend{YT-THDC,HAT}
    \end{axis}
  \end{tikzpicture}
  \vspace{-10pt}
  \caption{
    \textbf{Auxiliary-language oracle-translation diagnostic.}
    CER reductions are relative to the context-only baseline; bars below zero are better.
    Non-Mandarin auxiliary translations are produced by MT from oracle Mandarin translations.
  }
  \label{fig:auxiliary_language_delta}
  \vspace{-10pt}
\end{figure}

Figure \ref{fig:auxiliary_language_delta} shows that semantic anchoring is not limited to Mandarin: all MT-derived auxiliary languages outperform the context-only baseline, including typologically distant languages such as English, Spanish, and French.
Because these MT-derived auxiliary translations are translated from oracle Mandarin translations, the auxiliary-language experiment is a controlled diagnostic of cross-lingual semantic transfer rather than a deployment-time claim about arbitrary automatic auxiliary translations.
Appendix \ref{app:typological_and_multilingual} further compares these trends with URIEL \cite{littell2017} typological distances.

\begin{table}[t]
  \centering
  \small
  \setlength{\tabcolsep}{3pt}
  \begin{tabular*}{0.98\columnwidth}{@{\extracolsep{\fill}}lccc cc@{}}
    \toprule
    \textbf{Text Encoder} & \textbf{Train} & \textbf{Eval} & \textbf{Aud.} & \textbf{YT-THDC} & \textbf{HAT} \\
    \midrule
    Random Init. & Oracle & Auto & Yes & 27.39 & 27.00 \\
    Whisper Emb. & Oracle & Auto & Yes & 33.91 & 28.13 \\
    \textbf{mBERT} & Oracle & Auto & Yes & \textbf{23.48} & \textbf{22.48} \\
    \bottomrule
  \end{tabular*}
  \caption{
    \textbf{Text-encoder ablation.}
    CERs (\%) in audio-enhanced SAMA-ASR with oracle training anchors and automatic evaluation anchors.
  }
  \label{tab:text_encoder_comparison}
  \vspace{-5pt}
\end{table}

\section{Conclusion}

We present SAMA-ASR, a lightweight decoder-side adapter that treats low-resource ASR as both an acoustic adaptation problem and a target-side generation problem.
SAMA-ASR conditions a frozen encoder--decoder ASR backbone on translation-derived semantic anchors while retaining an explicit acoustic anchor.
Experiments on Taiwanese Hokkien and Hakka show consistent gains over acoustic and prior prompt-based baselines and a strong semantic-only TG-ASR baseline, as well as further improvements when combined with LoRA.
Analyses indicate that reliable semantic anchors help early autoregressive decoding and that compact upstream ST models can provide useful anchors without oracle translations at test time.
Overall, acoustic-grounded semantic anchoring offers a practical way to exploit paired speech--translation data when target-language transcripts are scarce.

\newpage

\section*{Limitations}

While SAMA-ASR demonstrates promising results, we acknowledge several limitations.
First, the practical pipeline requires an upstream ST pass, semantic encoding, and SAMA-ASR decoding, which increases inference cost compared with context-only or audio-only baselines; Appendix \ref{app:inference_efficiency} quantifies the resulting latency, model-size, and memory trade-offs.
Although the measured real-time factor remains below 1.0 in our offline setting, future work should optimize the latency--accuracy trade-off through model distillation, shared encoders, or anchor caching for more latency-sensitive deployment.
Second, SAMA-ASR depends on the quality of the automatic semantic anchors.
Our translator-capacity and paired-data-budget analyses show that useful anchors need not be oracle translations, but they must remain sufficiently meaning-compatible; in extremely low-data or severely hallucinated settings, noisy anchors can reduce the benefit or mislead the decoder.
Settings without paired auxiliary translations require an alternative source of semantic supervision and are outside the scope of the current framework.
This motivates future work on confidence estimation, anchor filtering, and joint training strategies that reduce train--test translation-source mismatch.
Finally, our empirical validation is limited to two low-resource Sinitic target varieties, Taiwanese Hokkien and Hakka.
The non-Mandarin auxiliary-language and multilingual-anchor analyses are diagnostic because they use MT-derived translations from oracle Mandarin rather than independent deployment-time ST systems.
Generalizing acoustic-grounded semantic anchoring to typologically diverse targets, especially languages with agglutinative or polysynthetic morphologies, requires broader multilingual evaluation.

\section*{Acknowledgments}

This work was supported by the National Science and Technology Council of Taiwan under Grants NSTC 112-2221-E-001-009-MY3 and NSTC 115-2634-F-001-006.


\bibliography{references.bib}

@inproceedings{alayrac2022,
  title = {Flamingo: {{A}} Visual Language Model for Few-Shot Learning},
  booktitle = {Proc. {{NeurlPS}}},
  author = {Alayrac, Jean-Baptiste and Donahue, Jeff and Luc, Pauline and Miech, Antoine and Barr, Iain and Hasson, Yana and Lenc, Karel and Mensch, Arthur and Millican, Katie and Reynolds, Malcolm and Ring, Roman and Rutherford, Eliza and Cabi, Serkan and Han, Tengda and Gong, Zhitao and Samangooei, Sina and Monteiro, Marianne and Menick, Jacob and Borgeaud, Sebastian and Brock, Andrew and Nematzadeh, Aida and Sharifzadeh, Sahand and Binkowski, Mikolaj and Barreira, Ricardo and Vinyals, Oriol and Zisserman, Andrew and Simonyan, Karen},
  year = 2022
}

@inproceedings{arora2022,
  title = {Why Exposure Bias Matters: An Imitation Learning Perspective of Error Accumulation in Language Generation},
  booktitle = {Findings of {{ACL}}},
  author = {Arora, Kushal and El Asri, Layla and Bahuleyan, Hareesh and Cheung, Jackie},
  editor = {Muresan, Smaranda and Nakov, Preslav and Villavicencio, Aline},
  year = 2022
}

@inproceedings{bapna2022,
  title = {{{mSLAM}}: Massively Multilingual Joint Pre-Training for Speech and Text},
  booktitle = {Arxiv Preprint {{arXiv}}:2202.01374},
  author = {Bapna, Ankur and Cherry, Colin and Zhang, Yu and Jia, Ye and Johnson, Melvin and Cheng, Yong and Khanuja, Simran and Riesa, Jason and Conneau, Alexis},
  year = 2022,
  eprint = {2202.01374},
  archiveprefix = {arXiv}
}

@inproceedings{barrault2023,
  title = {{{SeamlessM4T}}: {{Massively}} Multilingual \& Multimodal Machine Translation},
  booktitle = {Arxiv Preprint {{arXiv}}:2308.11596},
  author = {Communication, Seamless and Barrault, Lo{\"i}c and Chung, Yu-An and Meglioli, Mariano Cora and Dale, David and Dong, Ning and Duquenne, Paul-Ambroise and Elsahar, Hady and Gong, Hongyu and Heffernan, Kevin and Hoffman, John and Klaiber, Christopher and Li, Pengwei and Licht, Daniel and Maillard, Jean and Rakotoarison, Alice and Sadagopan, Kaushik Ram and Wenzek, Guillaume and Ye, Ethan and Akula, Bapi and Chen, Peng-Jen and Hachem, Naji El and Ellis, Brian and Gonzalez, Gabriel Mejia and Haaheim, Justin and Hansanti, Prangthip and Howes, Russ and Huang, Bernie and Hwang, Min-Jae and Inaguma, Hirofumi and Jain, Somya and Kalbassi, Elahe and Kallet, Amanda and Kulikov, Ilia and Lam, Janice and Li, Daniel and Ma, Xutai and Mavlyutov, Ruslan and Peloquin, Benjamin and Ramadan, Mohamed and Ramakrishnan, Abinesh and Sun, Anna and Tran, Kevin and Tran, Tuan and Tufanov, Igor and Vogeti, Vish and Wood, Carleigh and Yang, Yilin and Yu, Bokai and Andrews, Pierre and Balioglu, Can and {Costa-juss{\`a}}, Marta R. and Celebi, Onur and Elbayad, Maha and Gao, Cynthia and Guzm{\'a}n, Francisco and Kao, Justine and Lee, Ann and Mourachko, Alexandre and Pino, Juan and Popuri, Sravya and Ropers, Christophe and Saleem, Safiyyah and Schwenk, Holger and Tomasello, Paden and Wang, Changhan and Wang, Jeff and Wang, Skyler},
  year = 2023,
  eprint = {2308.11596},
  archiveprefix = {arXiv}
}

@inproceedings{bengio2015,
  title = {Scheduled Sampling for Sequence Prediction with Recurrent Neural Networks},
  booktitle = {Proc. {{NeurIPS}}},
  author = {Bengio, Samy and Vinyals, Oriol and Jaitly, Navdeep and Shazeer, Noam},
  year = 2015
}

@article{besacier2014,
  title = {Automatic Speech Recognition for Under-Resourced Languages: {{A}} Survey},
  author = {Besacier, Laurent and Barnard, Etienne and Karpov, Alexey and Schultz, Tanja},
  year = 2014,
  journal = {Speech Communication},
  volume = {56},
  pages = {85--100}
}

@inproceedings{bisani2004,
  title = {Bootstrap Estimates for Confidence Intervals in {{ASR}} Performance Evaluation},
  booktitle = {Proc. {{ICASSP}}},
  author = {Bisani, M. and Ney, H.},
  year = 2004
}

@inproceedings{chen2020,
  title = {Using Taigi Dramas with Mandarin Chinese Subtitles to Improve Taigi Speech Recognition},
  booktitle = {Proc. {{O-COCOSDA}}},
  author = {Chen, Pin-Yuan and Wu, Chia-Hua and Lee, Hung-Shin and Tsao, Shao-Kang and Ko, Ming-Tat and Wang, Hsin-Min},
  year = 2020
}

@inproceedings{chengExploringImpactData2025,
  title = {Exploring the Impact of Data Quantity on {{ASR}} in Extremely Low-Resource Languages},
  booktitle = {Proc. {{O-COCOSDA}}},
  author = {Cheng, Yao-Fei and Chen, Li-Wei and Lee, Hung-Shin and Wang, Hsin-Min},
  year = 2025
}

@inproceedings{chenIts2024,
  title = {It's Never Too Late: {{Fusing}} Acoustic {{Information}} into Large Language Models for Automatic Speech Recognition},
  booktitle = {Proc. {{ICLR}}},
  author = {Chen, Chen and Li, Ruizhe and Hu, Yuchen and Siniscalchi, Sabato and Chen, Pin-Yu and Chng, Ensiong and Yang, Chao-Han Huck},
  year = 2024
}

@inproceedings{choi2022,
  title = {Distilling a Pretrained Language Model to a Multilingual {{ASR}} Model},
  booktitle = {Proc. {{Interspeech}}},
  author = {Choi, Kwanghee and Park, Hyung-Min},
  year = 2022
}

@inproceedings{conneau2021,
  title = {Unsupervised Cross-Lingual Representation Learning for Speech Recognition},
  booktitle = {Proc. {{Interspeech}}},
  author = {Conneau, Alexis and Baevski, Alexei and Collobert, Ronan and Mohamed, Abdelrahman and Auli, Michael},
  year = 2021
}

@inproceedings{coto-solano2022,
  title = {Evaluating Word Embeddings in Extremely Under-Resourced Languages: {{A}} Case Study in {{Bribri}}},
  booktitle = {Proc. {{COLING}}},
  author = {{Coto-Solano}, Rolando},
  editor = {Calzolari, Nicoletta and Huang, Chu-Ren and Kim, Hansaem and Pustejovsky, James and Wanner, Leo and Choi, Key-Sun and Ryu, Pum-Mo and Chen, Hsin-Hsi and Donatelli, Lucia and Ji, Heng and Kurohashi, Sadao and Paggio, Patrizia and Xue, Nianwen and Kim, Seokhwan and Hahm, Younggyun and He, Zhong and Lee, Tony Kyungil and Santus, Enrico and Bond, Francis and Na, Seung-Hoon},
  year = 2022
}

@inproceedings{devlin2019,
  title = {{{BERT}}: {{Pre-training}} of Deep Bidirectional Transformers for Language Understanding},
  booktitle = {Proc. {{NAACL}}},
  author = {Devlin, Jacob and Chang, Ming-Wei and Lee, Kenton and Toutanova, Kristina},
  year = 2019
}

@inproceedings{feng2022,
  title = {Language-Agnostic {{BERT}} Sentence Embedding},
  booktitle = {Proc. {{ACL}}},
  author = {Feng, Fangxiaoyu and Yang, Yinfei and Cer, Daniel and Arivazhagan, Naveen and Wang, Wei},
  year = 2022
}

@inproceedings{getman2024,
  title = {Exploring Adaptation Techniques of Large Speech Foundation Models for Low-Resource {{ASR}}: A Case Study on {{Northern S\'ami}}},
  booktitle = {Proc. {{Interspeech}}},
  author = {Getman, Yaroslav and Grosz, Tamas and {Hiovain-Asikainen}, Katri and Kurimo, Mikko},
  year = 2024
}

@inproceedings{hentschel2024,
  title = {Keep Decoding Parallel with Effective Knowledge Distillation from Language Models to End-to-End Speech Recognisers},
  booktitle = {Proc. {{ICASSP}}},
  author = {Hentschel, Michael and Nishikawa, Yuta and Komatsu, Tatsuya and Fujita, Yusuke},
  year = 2024
}

@inproceedings{hou2021,
  title = {Meta-Adapter: Efficient Cross-Lingual Adaptation with Meta-Learning},
  booktitle = {Proc. {{ICASSP}}},
  author = {Hou, Wenxin and Wang, Yidong and Gao, Shengzhou and Shinozaki, Takahiro},
  year = 2021
}

@inproceedings{hsu2025,
  title = {Let's Fuse Step by Step: {{A}} Generative Fusion Decoding Algorithm with {{LLMs}} for Robust and Instruction-Aware {{ASR}} and {{OCR}}},
  booktitle = {Findings of {{ACL}}},
  author = {Hsu, Chan-Jan and Chen, Yi-Chang and Liao, Feng-Ting and Ho, Pei-Chen and Wang, Yu-Hsiang and Hsu, Po-Chun and Shiu, Da-shan},
  editor = {Che, Wanxiang and Nabende, Joyce and Shutova, Ekaterina and Pilehvar, Mohammad Taher},
  year = 2025
}

@inproceedings{ko2015,
  title = {Audio Augmentation for Speech Recognition},
  booktitle = {Proc. {{Interspeech}}},
  author = {Ko, Tom and Peddinti, Vijayaditya and Povey, Daniel and Khudanpur, Sanjeev},
  year = 2015
}

@inproceedings{le2021,
  title = {Contextualized Streaming End-to-End Speech Recognition with Trie-Based Deep Biasing and Shallow Fusion},
  booktitle = {Proc. {{Interspeech}}},
  author = {Le, Duc and Jain, Mahaveer and Keren, Gil and Kim, Suyoun and Shi, Yangyang and Mahadeokar, Jay and Chan, Julian and Shangguan, Yuan and Fuegen, Christian and Kalinli, Ozlem and Saraf, Yatharth and Seltzer, Michael L.},
  year = 2021
}

@inproceedings{leDualdecoderTransformerJoint2020,
  title = {Dual-Decoder Transformer for Joint Automatic Speech Recognition and Multilingual Speech Translation},
  booktitle = {Proc. {{COLING}}},
  author = {Le, Hang and Pino, Juan and Wang, Changhan and Gu, Jiatao and Schwab, Didier and Besacier, Laurent},
  editor = {Scott, Donia and Bel, Nuria and Zong, Chengqing},
  year = 2020
}

@inproceedings{li2026,
  title = {Multimodal In-Context Learning for {{ASR}} of Low-Resource Languages},
  booktitle = {Findings of {{ACL}}},
  author = {Li, Zhaolin and Niehues, Jan},
  year = 2026
}

@inproceedings{liao2023,
  title = {Taiwanese {{Hakka}} across {{Taiwan}} Corpus and {{Formosa}} Speech Recognition Challenge 2023 - {{Hakka ASR}}},
  booktitle = {Proc. {{O-COCOSDA}}},
  author = {Liao, Yuan-Fu and Hwang, Shaw-Hwa and Chen, You-Shuo and Lai, Han-Chun and Chung, Yao-Hsing and Shen, Li-Te and Huang, Yen-Chun and Huang, Chi-Jung and Han, Hsu Wen and Chen, Li-Wei and Su, Pei-Chung and Huang, Chao-Shih},
  year = 2023
}

@inproceedings{littell2017,
  title = {{{URIEL}} and Lang2vec: Representing Languages as Typological, Geographical, and Phylogenetic Vectors},
  booktitle = {Proc. {{EACL}}},
  author = {Littell, Patrick and Mortensen, David R. and Lin, Ke and Kairis, Katherine and Turner, Carlisle and Levin, Lori},
  editor = {Lapata, Mirella and Blunsom, Phil and Koller, Alexander},
  year = 2017
}

@inproceedings{loshchilov2019,
  title = {Decoupled Weight Decay Regularization},
  booktitle = {Proc. {{ICLR}}},
  author = {Loshchilov, Ilya and Hutter, Frank},
  year = 2019
}

@inproceedings{ma2021,
  title = {End-to-End Audio-Visual Speech Recognition with Conformers},
  booktitle = {Proc. {{ICASSP}}},
  author = {Ma, Pingchuan and Petridis, Stavros and Pantic, Maja},
  year = 2021
}

@inproceedings{park2019,
  title = {{{SpecAugment}}: {{A}} Simple Data Augmentation Method for Automatic Speech Recognition},
  booktitle = {Proc. {{Interspeech}}},
  author = {Park, Daniel S. and Chan, William and Zhang, Yu and Chiu, Chung-Cheng and Zoph, Barret and Cubuk, Ekin D. and Le, Quoc V.},
  year = 2019
}

@inproceedings{peng2023,
  title = {Prompting the Hidden Talent of Web-Scale Speech Models for Zero-Shot Task Generalization},
  booktitle = {Proc. {{Interspeech}}},
  author = {Peng, Puyuan and Yan, Brian and Watanabe, Shinji and Harwath, David},
  year = 2023
}

@inproceedings{peng2026,
  title = {Efficient Dialect-Aware Modeling and Conditioning for Low-Resource {{Taiwanese Hakka}} Speech Processing},
  booktitle = {Proc. {{LREC}}},
  author = {Peng, An-Ci and Huang, Kuan-Tang and Lo, Tien-Hong and Lee, Hung-Shin and Wang, Hsin-Min and Chen, Berlin},
  year = 2026
}

@inproceedings{pluss2023,
  title = {{{STT4SG-350}}: {{A}} Speech Corpus for All {{Swiss German}} Dialect Regions},
  booktitle = {Proc. {{ACL}}},
  author = {Pl{\"u}ss, Michel and Deriu, Jan and Schraner, Yanick and Paonessa, Claudio and Hartmann, Julia and Schmidt, Larissa and Scheller, Christian and H{\"u}rlimann, Manuela and Samard{\v z}i{\'c}, Tanja and Vogel, Manfred and Cieliebak, Mark},
  editor = {Rogers, Anna and {Boyd-Graber}, Jordan and Okazaki, Naoaki},
  year = 2023
}

@inproceedings{rouditchenko2024,
  title = {Whisper-Flamingo: {{Integrating}} Visual Features into Whisper for Audio-Visual Speech Recognition and Translation},
  booktitle = {Proc. {{Interspeech}}},
  author = {Rouditchenko, Andrew and Gong, Yuan and Thomas, Samuel and Karlinsky, Leonid and Kuehne, Hilde and Feris, Rogerio and Glass, James},
  year = 2024
}

@inproceedings{songLoRAWhisperParameterEfficientExtensible2024,
  title = {{{LoRA-Whisper}}: {{Parameter-efficient}} and Extensible Multilingual {{ASR}}},
  booktitle = {Proc. {{Interspeech}}},
  author = {Song, Zheshu and Zhuo, Jianheng and Yang, Yifan and Ma, Ziyang and Zhang, Shixiong and Chen, Xie},
  year = 2024
}

@inproceedings{sterpu2018,
  title = {Attention-Based Audio-Visual Fusion for Robust Automatic Speech Recognition},
  booktitle = {Proc. {{ICMI}}},
  author = {Sterpu, George and Saam, Christian and Harte, Naomi},
  year = 2018
}

@inproceedings{sun2023,
  title = {Can Contextual Biasing Remain Effective with Whisper and {{GPT-2}}?},
  booktitle = {Proc. {{Interspeech}}},
  author = {Sun, Guangzhi and Zheng, Xianrui and Zhang, Chao and Woodland, Philip C.},
  year = 2023
}

@inproceedings{taniguchi2022,
  title = {Transformer-Based Automatic Speech Recognition with Auxiliary Input of Source Language Text toward Transcribing Simultaneous Interpretation},
  booktitle = {Proc. {{Interspeech}}},
  author = {Taniguchi, Shuta and Kato, Tsuneo and Tamura, Akihiro and Yasuda, Keiji},
  year = 2022
}

@article{watanabe2017,
  title = {Hybrid {{CTC}}/Attention Architecture for End-to-End Speech Recognition},
  author = {Watanabe, Shinji and Hori, Takaaki and Kim, Suyoun and Hershey, John R. and Hayashi, Tomoki},
  year = 2017,
  journal = {IEEE Journal of Selected Topics in Signal Processing},
  volume = {11},
  number = {8},
  pages = {1240--1253}
}

@inproceedings{xiao2023,
  title = {{{HK-LegiCoST}}: {{Leveraging}} Non-Verbatim Transcripts for Speech Translation},
  booktitle = {Proc. {{Interspeech}}},
  author = {Xiao, Cihan and Xinyuan, Henry Li and Yang, Jinyi and Gao, Dongji and Wiesner, Matthew and Duh, Kevin and Khudanpur, Sanjeev},
  year = 2023
}

@inproceedings{yang2025,
  title = {Enhancing Low-Resource {{ASR}} through Versatile {{TTS}}: Bridging the Data Gap},
  booktitle = {Proc. {{ICASSP}}},
  author = {Yang, Guanrou and Yu, Fan and Ma, Ziyang and Du, Zhihao and Gao, Zhifu and Zhang, Shiliang and Chen, Xie},
  year = 2025
}

@inproceedings{yang2026,
  title = {{{TG-ASR}}: {{Translation-guided}} Learning with Parallel Gated Cross Attention for Low-Resource Automatic Speech Recognition},
  booktitle = {Proc. {{LREC}}},
  author = {Yang, Cheng-Yeh and Wang, Chien-Chun and Chen, Li-Wei and Lee, Hung-Shin and Wang, Hsin-Min and Chen, Berlin},
  year = 2026
}

@inproceedings{yeroyan2024,
  title = {Enabling {{ASR}} for Low-Resource Languages: A Comprehensive Dataset Creation Approach},
  booktitle = {Arxiv Preprint {{arXiv}}:2406.01446},
  author = {Yeroyan, Ara and Karpov, Nikolay},
  year = 2024,
  eprint = {2406.01446},
  archiveprefix = {arXiv}
}

@inproceedings{zanonboito2022,
  title = {Speech {{Resources}} in the {{Tamasheq Language}}},
  booktitle = {Proc. {{LREC}}},
  author = {Zanon Boito, Marcely and Bougares, Fethi and Barbier, Florentin and Gahbiche, Souhir and Barrault, Lo{\"i}c and Rouvier, Mickael and Est{\`e}ve, Yannick},
  editor = {Calzolari, Nicoletta and B{\'e}chet, Fr{\'e}d{\'e}ric and Blache, Philippe and Choukri, Khalid and Cieri, Christopher and Declerck, Thierry and Goggi, Sara and Isahara, Hitoshi and Maegaard, Bente and Mariani, Joseph and Mazo, H{\'e}l{\`e}ne and Odijk, Jan and Piperidis, Stelios},
  year = 2022
}

@inproceedings{zhao2019,
  title = {Shallow-Fusion End-to-End Contextual Biasing},
  booktitle = {Proc. {{Interspeech}}},
  author = {Zhao, Ding and Sainath, Tara N. and Rybach, David and Rondon, Pat and Bhatia, Deepti and Li, Bo and Pang, Ruoming},
  year = 2019
}

\clearpage
\appendix

\section{Additional Experimental Details}
\label{app:experimental_details}

We confirm that all pre-trained models and datasets employed in our experiments are publicly available. Our usage of these existing artifacts is strictly consistent with their intended research purposes and complies with their respective academic licenses. Furthermore, the public benchmark datasets utilized in this study have been previously curated by their creators to exclude Personally Identifiable Information (PII) and offensive content. Any new artifacts or evaluation scripts generated from this study are carefully reviewed to ensure anonymity and are intended solely for non-commercial, academic research.

\paragraph{Dataset Construction.}
YT-THDC follows the previously introduced Taiwanese Hokkien split and preprocessing protocol \cite{yang2026}; its Mandarin translations are used as loosely aligned auxiliary translations rather than verbatim transcripts.
For Hakka, we use the Hakka Across Taiwan (HAT) corpus \cite{liao2023}.
Although the full HAT corpus contains approximately 600 hours covering Sixian and Hailu dialects, we construct a controlled low-resource subset by randomly sampling 30 hours of speech and restricting utterance durations to 2--10 seconds.
The controlled-subset design keeps the evaluation focused on low-resource conditions while preserving dialectal diversity.

\paragraph{Optimization Details.}
All ASR systems use the original pretrained Whisper tokenizer, which supports multilingual text including Chinese characters, so no external vocabulary mapping is required.
All models are optimized with AdamW \cite{loshchilov2019} using a learning rate of $1.0 \times 10^{-4}$, a batch size of 32, and a weight decay of 0.01.
We train for 32k steps on YT-THDC and 20k steps on HAT to account for differences in utterance count.
A linear learning-rate scheduler with a 10\% warm-up phase is applied.
Audio inputs are truncated or padded to 10~seconds and converted into 80-channel log-mel spectrograms.
For SAMA-ASR, both acoustic and cross-lingual text embeddings are projected to dimension $D=1024$.

\paragraph{Prompt Baseline Decoding.}
Self Attn.+Prompt uses the same automatic Mandarin translations as the practical TG-ASR and SAMA-ASR rows.
For each test utterance, we remove Whisper control tokens from the generated translation and pass the remaining Mandarin string to Whisper's native \texttt{prompt} field in \texttt{DecodingOptions}.
Prompted decoding uses \texttt{task=transcribe}, \texttt{language=zh}, no timestamps, beam size 1, and temperature 0; because Whisper prompts are specified per decoding option, prompted utterances are decoded one at a time and the prompt is reset for each utterance.

\paragraph{Model Size and Computational Budget.}
We implemented all experiments using PyTorch. The complete architecture contains approximately 1.3B parameters in total, of which roughly 404M are trainable parameters during our training phase.
Unless otherwise specified, all experiments use a single NVIDIA RTX 3090 (24GB) GPU.
The computational budget for training a single experimental run is estimated to be approximately 12 GPU hours.

\paragraph{AI Assistant Usage.}
During the preparation of this manuscript, we used Prism AI and Gemini to assist with language polishing, grammar correction, and code generation.
We explicitly state that all AI-generated text and code were strictly reviewed, verified, and thoroughly tested by the human authors.
These tools were not used to make autonomous decisions about methodology, evaluation, or interpretation of results.
No generative AI was used to conceive novel scientific ideas, experimental designs, or draw conclusions. The authors assume full responsibility for the accuracy, originality, and integrity of the final manuscript and the implemented codebase.

\begin{table}[!t]
  \small
  \centering
  \setlength{\tabcolsep}{7pt}
  \begin{tabular}{lcccc}
    \toprule
    \multirow{2}{*}{\textbf{Split}} & \multicolumn{2}{c}{\textbf{YT-THDC}} & \multicolumn{2}{c}{\textbf{HAT}} \\
    \cmidrule(lr){2-3} \cmidrule(lr){4-5}
    & \textbf{Duration} & \textbf{\# Utts.} & \textbf{Duration} & \textbf{\# Utts.} \\
    \midrule
    Train & 27.51 & 50,984 & 27.50 & 19,120 \\
    Test & 2.79 & 4,859 & 2.50 & 1,725 \\
    \midrule
    Total & 30.30 & 55,843 & 30.00 & 20,845 \\
    \bottomrule
  \end{tabular}
  \vspace{-5pt}
  \caption{
    \textbf{Low-resource dataset statistics.}
    Durations are in hours.
  }
  \label{tab:dataset_statistic}
\end{table}

\section{Robustness under Limited Training Data and Noisy Translation Resources}
\label{app:training_anchor_source}

\paragraph{Limited Training Data.}
Table \ref{tab:training_anchor_source} addresses the train--test mismatch discussed in Section \ref{ssec:training_objective}: main experiments train SAMA-ASR with oracle translations but evaluate practical rows with automatic translations.
Pseudo-translation training matches inference more closely by using translator-generated translations.

\begin{table}[t]
  \centering
  \small
  \setlength{\tabcolsep}{4pt}
  \begin{tabular}{llcc}
    \toprule
    \textbf{Train Set} & \textbf{Train Translation} & \textbf{YT-THDC} & \textbf{HAT} \\
    \midrule
    30 hr & Pseudo & 23.93 & \textbf{21.51} \\
    30 hr & Oracle & \textbf{23.48} & 22.48 \\
    \midrule
    10 hr & Pseudo & 50.50 & 39.53 \\
    10 hr & Oracle & \textbf{30.04} & \textbf{30.93} \\
    \midrule
    1 hr & Pseudo & 56.29 & 65.97 \\
    1 hr & Oracle & \textbf{53.45} & \textbf{54.68} \\
    \midrule
    10 min & Pseudo & 67.16 & 81.19 \\
    10 min & Oracle & \textbf{65.63} & \textbf{72.67} \\
    \bottomrule
  \end{tabular}
  \caption{
    \textbf{Training-translation source ablation.}
    All rows evaluate with automatic Mandarin auxiliary translations; values are CERs (\%).
    Pseudo-translation training uses translator-generated auxiliary translations, while oracle-translation training uses paired Mandarin translations.
  }
  \label{tab:training_anchor_source}
\end{table}

Under the full 30-hour setting, the two strategies are close and mixed: oracle-translation training is slightly better on YT-THDC, while pseudo-translation training is better on HAT.
The 30-hour ablation suggests that the train--test translation-source mismatch is not large when enough paired data is available to train a usable translator.
When paired data is reduced, however, oracle-translation training consistently outperforms pseudo-translation training because the translator-generated translations become much noisier.
We therefore use oracle-translation training in the main experiments to avoid propagating translator errors into SAMA-ASR, while keeping evaluation translations automatic for practical deployment.

{
  \paragraph{Noisy Translation Resources.}
  We further consider a separate setting where the full 30-hour training set is retained, but both SAMA-ASR training and evaluation use pseudo translations generated by ST models of different capacities.

  \begin{table}[t]
    \centering
    \small
    \setlength{\tabcolsep}{5pt}
    \begin{tabular}{lcc}
      \toprule
      \textbf{System / ST Source} & \textbf{BLEU} $\uparrow$ & \textbf{CER} $\downarrow$ \\
      \midrule
      Cross Attn. baseline & -- & 32.89 \\
      SAMA-ASR + Tiny & 32.37 & 32.56 \\
      SAMA-ASR + Base & 39.03 & 31.90 \\
      SAMA-ASR + Small & 52.39 & 26.80 \\
      SAMA-ASR + Medium & 54.75 & 24.74 \\
      SAMA-ASR + Large & 59.57 & 24.14 \\
      \bottomrule
    \end{tabular}
    \caption{
      \textbf{Robustness to noisy translation resources on YT-THDC.}
      Both training and evaluation use pseudo Mandarin translations generated by the indicated ST model; values are BLEU and CER (\%).
    }
    \label{tab:noisy_translation_resources}
  \end{table}

  Table \ref{tab:noisy_translation_resources} shows that SAMA-ASR remains effective when semantic anchors are automatically generated throughout both training and evaluation.
  Even Tiny provides a small improvement over the Cross Attn. baseline, while stronger ST models progressively reduce CER as translation quality improves.
  Together with the limited-data analysis above, these results distinguish two sources of difficulty: reducing paired data weakens both adaptation and translation, whereas this setting isolates translation noise while retaining the full training set.
}

\section{Semantic-Neighborhood Visualization}
\label{app:semantic_neighborhood_visualization}

\begin{figure}[t]
  \centering
  \includegraphics[width=\linewidth]{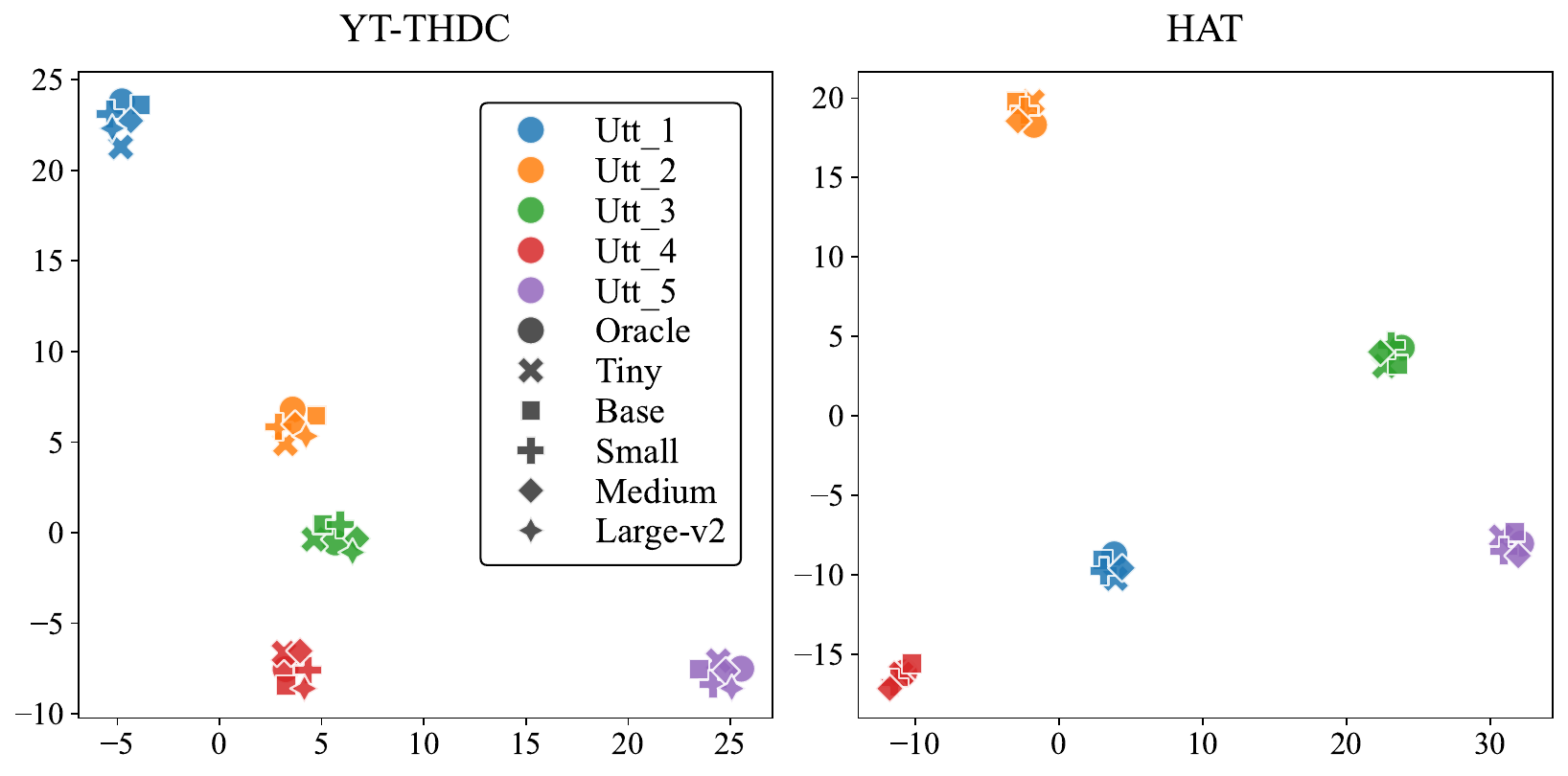}
  \vspace{-8pt}
  \caption{
    \textbf{Semantic-neighborhood visualization of automatic translations.}
    UMAP shows mBERT embeddings for five sampled test utterances per dataset.
    Colors denote utterances, marker shapes denote translation sources, and axes are for visualization only.
  }
  \label{fig:semantic_umap}
  \vspace{-10pt}
\end{figure}

To further examine why automatic semantic anchors can remain useful despite imperfect translation quality, we visualize the mBERT embedding neighborhoods of oracle and automatic Mandarin translations.
Figure \ref{fig:semantic_umap} shows UMAP projections for five sampled test utterances from each dataset.
Automatic translations for the same utterance tend to remain close to the corresponding oracle translation, supporting the interpretation from the translator-capacity analysis in Figure \ref{fig:scalability_analysis} that approximate semantic-neighborhood compatibility can guide decoding when the generated anchor remains meaning-compatible.

\section{Detailed Case Study}
\label{sec:appendix_case_study}

To qualitatively inspect the error patterns behind the observed gains, Table \ref{tab:case_study_yt_thdc} presents a representative YT-THDC example in which a short target phrase is acoustically confusable with several semantically incompatible alternatives.

\begin{table*}[htbp]
  \centering
  \small
  \begin{CJK*}{UTF8}{bsmi} 
    \begin{tabular}{@{}ll@{}}
      \toprule
      \textbf{System} & \textbf{Transcription and English Translation (Gloss)} \\
      \midrule
      \textbf{REF} & 干焦會當靠\textcolor{red}{到銀行}抑是公家機關看診 \\
      & \textit{(Can only rely on \textcolor{red}{arriving at banks} or government agencies for medical consultations.)} \\
      \midrule
      \textbf{Self Attn.} & 干焦會當靠\textcolor{red}{教囡仔}抑是公家機關看證 \\
      & \textit{(Can only rely on \textcolor{red}{teaching children} or government agencies for verification.)} \\
      \midrule
      \textbf{Cross Attn.} & 就按呢咱靠\textcolor{red}{教議員學}抑是公家機關看證 \\
      & \textit{(Just like this, we rely on \textcolor{red}{teaching councilors to learn} or government agencies for verification.)} \\
      \midrule
      \textbf{TG-ASR} & 干焦會當靠\textcolor{red}{交警方}抑是公家機關看證 \\
      & \textit{(Can only rely on \textcolor{red}{turning over to the police} or government agencies for verification.)} \\
      \midrule
      \textbf{SAMA-ASR} & 干焦會當靠\textcolor{red}{到銀行}抑是公家機關看證 \\
      & \textit{(Can only rely on \textcolor{red}{arriving at banks} or government agencies for verification.)} \\
      \bottomrule
    \end{tabular}

    \medskip 

    \caption[An error correction case study comparing baseline models and SAMA-ASR on the YT-THDC dataset.]{\textbf{YT-THDC error correction case study.} Red highlights mark the target phrase and corresponding substitutions. Baselines produce acoustically or semantically plausible substitutions, whereas SAMA-ASR recovers 「到銀行」 (\textit{kàu gîn-hâng}, arriving at banks) by combining semantic guidance with acoustic grounding.}
    \label{tab:case_study_yt_thdc}
  \end{CJK*} 
\end{table*}

As shown in Table~\ref{tab:case_study_yt_thdc}, the target sequence
\begin{CJK*}{UTF8}{bsmi}「到銀行」
\end{CJK*} (\textit{kàu gîn-hâng})
is a difficult case because it can be confused with phonetically similar expressions in Taiwanese Hokkien.
The two baselines without semantic anchoring, \textbf{Self Attn.} and \textbf{Cross Attn.}, therefore make pronunciation-driven substitutions.
\textbf{Self Attn.} replaces the target phrase with
\begin{CJK*}{UTF8}{bsmi}「教囡仔」
\end{CJK*} (\textit{kà gín-á}, meaning ``teaching children''), while \textbf{Cross Attn.} maps the same target slot to
\begin{CJK*}{UTF8}{bsmi}「教議員學」
\end{CJK*}
(\textit{kà gī-uân h\textsuperscript{\textbar}k}, meaning ``teaching councilors to learn'').
Both outputs preserve local phonetic similarity, but they are semantically incompatible with the utterance.

By contrast, the text-guided model (\textbf{TG-ASR}) has semantic context, but lacks an explicit acoustic anchor. The nearby mention of ``government agencies''
(
  \begin{CJK*}{UTF8}{bsmi}公家機關
\end{CJK*})
encourages it to predict the semantically associated but acoustically unsupported phrase ``turning over to the police''
(
  \begin{CJK*}{UTF8}{bsmi}「交警方」
\end{CJK*}).

\textbf{SAMA-ASR} avoids both types of errors by using the two anchors for complementary roles. The semantic anchor rejects phonetically plausible but meaning-incompatible hypotheses such as
\begin{CJK*}{UTF8}{bsmi}「教囡仔」
\end{CJK*} and
\begin{CJK*}{UTF8}{bsmi}「教議員學」
\end{CJK*},
while the acoustic anchor prevents text-side hallucinations such as
\begin{CJK*}{UTF8}{bsmi}「交警方」
\end{CJK*}.
Consequently, the model recovers the correct phrase
\begin{CJK*}{UTF8}{bsmi}「到銀行」
\end{CJK*}, illustrating why semantic guidance and acoustic verification are most effective when used together.

\section{Typological and Multilingual Analyses}
\label{app:typological_and_multilingual}

\begin{table}[t]
  \small
  \centering
  \setlength{\tabcolsep}{7pt}
  \begin{tabular}{lccc}
    \toprule
    \textbf{Language} & \textbf{Syntactic} & \textbf{Phonological} & \textbf{Inventory} \\
    \midrule
    \multicolumn{4}{l}{\textit{\textbf{Distances from Taiwanese Hokkien}}} \\
    \multicolumn{4}{l}{\textit{\textbf{Sinitic Family}}} \\
    Mandarin & \textbf{0.58} & \textbf{0.53} & 0.65 \\
    \midrule
    \multicolumn{4}{l}{\textit{\textbf{Indo-European Family}}} \\
    English & 0.62 & 0.64 & \textbf{0.55} \\
    Hindi & 0.70 & 0.62 & 0.59 \\
    Spanish & 0.66 & 0.62 & 0.63 \\
    French & 0.68 & 0.65 & 0.57 \\
    \midrule
    \multicolumn{4}{l}{\textit{\textbf{Distances from Hakka}}} \\
    \multicolumn{4}{l}{\textit{\textbf{Sinitic Family}}} \\
    Mandarin & \textbf{0.52} & \textbf{0.41} & 0.55 \\
    \midrule
    \multicolumn{4}{l}{\textit{\textbf{Indo-European Family}}} \\
    English & 0.58 & 0.52 & \textbf{0.54} \\
    Hindi & 0.68 & 0.49 & 0.58 \\
    Spanish & 0.64 & 0.49 & 0.56 \\
    French & 0.71 & 0.55 & 0.55 \\
    \bottomrule
  \end{tabular}
  \vspace{-5pt}
  \caption{
    \textbf{URIEL linguistic distances.}
    Cosine distances are measured from Taiwanese Hokkien and Hakka; lower is closer.
  }
  \label{tab:linguistic_distance}
\end{table}

To investigate whether auxiliary-language performance is determined solely by linguistic similarity, we compare Figure \ref{fig:auxiliary_language_delta} with the URIEL typological distances in Table \ref{tab:linguistic_distance}.
Mandarin has the lowest syntactic and phonological distances to both Taiwanese Hokkien and Hakka, consistent with its strong performance as an oracle translation.
However, the non-Sinitic results do not follow a simple monotonic relationship with typological distance.
For example, French yields stronger gains than English on both datasets despite being syntactically and phonologically farther from the corresponding target variety.
The typological comparison supports our interpretation that SAMA-ASR operates in a continuous semantic space and can exploit cross-lingual conceptual information in this oracle-reference-derived diagnostic setting rather than relying only on surface-level similarity.

\begin{figure}[t]
  \centering
  \includegraphics[width=\linewidth]{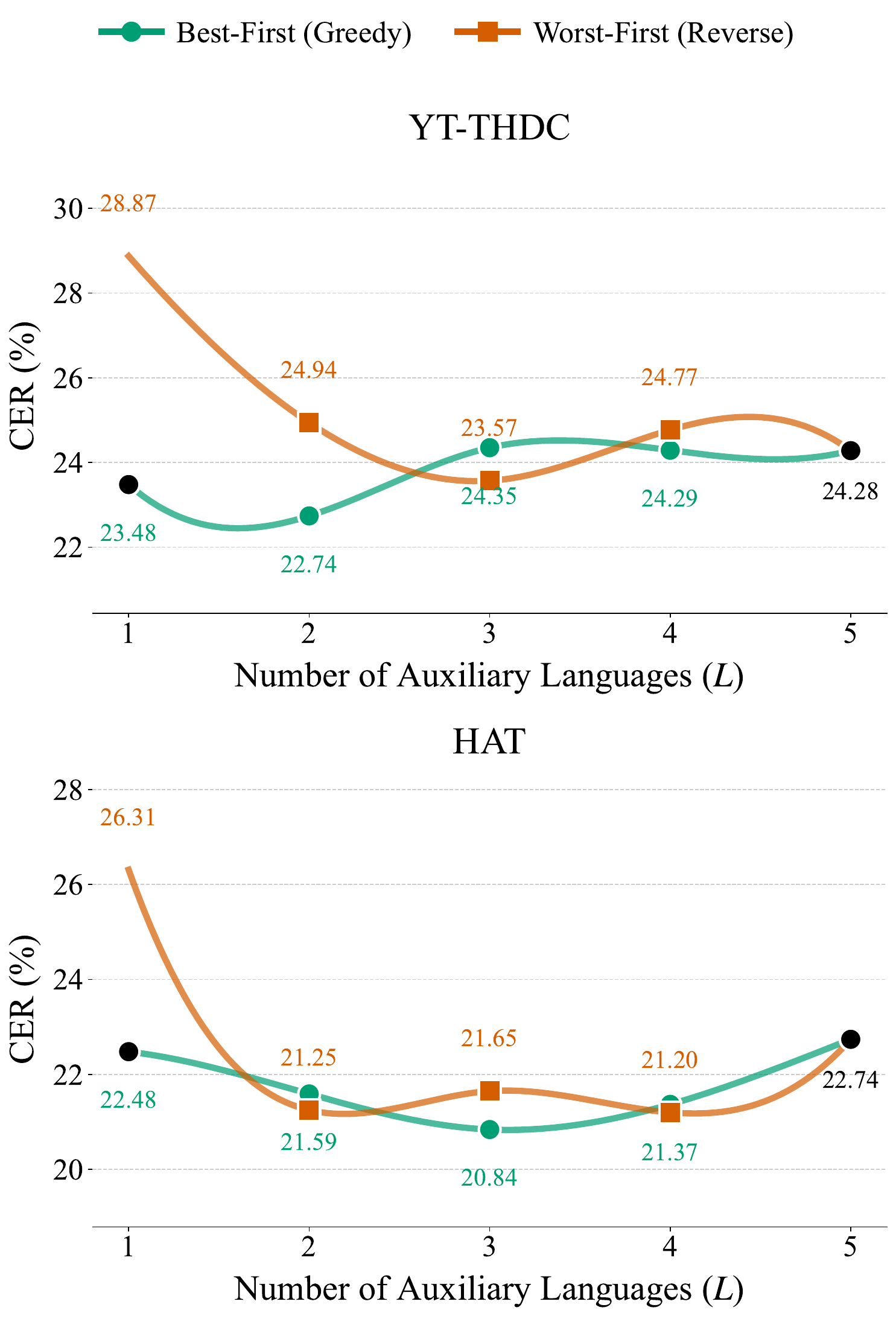}
  \vspace{-10pt}
  \caption{
    \textbf{Multilingual translation-composition diagnostic.}
    Best-first and worst-first orders add auxiliary translations by oracle single-language performance; $L$ denotes the number of auxiliary translations.
  }
  \label{fig:synergistic_effect}
\end{figure}

We further evaluate whether multiple auxiliary translations provide complementary semantic evidence.
As shown in Figure \ref{fig:synergistic_effect}, the benefit of multilingual anchoring depends strongly on translation composition.
The best-first strategy consistently outperforms the worst-first strategy, especially when only one additional language is introduced ($L=2$), indicating that carefully selected auxiliary translations provide useful complementary semantics beyond Mandarin.
As more languages are added, the trend becomes dataset-dependent: some settings saturate earlier, while Hakka remains competitive up to larger multilingual sets and degrades only when all five languages are included.
These results suggest that multilingual anchoring is not a simple ``more languages is better'' effect; rather, SAMA-ASR benefits from high-quality complementary translations, whereas redundant or less compatible translations can dilute the attention signal.

{
  \section{Inference Efficiency and Deployment Cost Analysis}
  \label{app:inference_efficiency}

  To quantify the accuracy--efficiency trade-off of semantic-anchor generation and acoustic grounding, we report real-time factor (RTF), total model parameters, and peak GPU memory on the full YT-THDC test set.
  RTF is defined as wall-clock processing time divided by input-audio duration.
  All efficiency measurements use batch size 1, exclude the first 10 warm-up utterances from timing, and follow the same per-utterance greedy decoding protocol as the main experiments.

  \begin{table}[t]
    \centering
    \small
    \setlength{\tabcolsep}{2.5pt}
    \resizebox{\columnwidth}{!}{%
      \begin{tabular}{lccrrrr}
        \toprule
        \textbf{Method} & \textbf{Sem.} & \textbf{Aud.} & \textbf{CER} $\downarrow$ & \textbf{RTF} $\downarrow$ & \textbf{Params (B)} & \textbf{VRAM (GB)} \\
        \midrule
        Self Attn. & No & No & 33.49 & 0.216 & 0.942 & 3.66 \\
        TG-ASR (Oracle) & Yes & No & 20.15 & 0.307 & 1.244 & 4.86 \\
        SAMA-ASR (Oracle) & Yes & Yes & 16.79 & 0.391 & 1.345 & 5.22 \\
        \bottomrule
      \end{tabular}%
    }
    \caption{
      \textbf{Inference cost with provided translations on YT-THDC.}
      Oracle translations are available at inference time in this controlled setting, isolating the overhead of semantic and acoustic anchoring from upstream ST generation.
      ``Sem.'' and ``Aud.'' indicate the explicit semantic- and acoustic-anchor branches.
    }
    \label{tab:controlled_inference_cost}
  \end{table}

  Table \ref{tab:controlled_inference_cost} first isolates the cost of the anchoring modules by assuming that translations are already available.
  Compared with TG-ASR, SAMA-ASR provides a 16.7\% relative CER reduction while increasing the total parameter count by 8.1\% and peak GPU memory by 7.4\%.
  Its RTF increases from 0.307 to 0.391, but remains well below 1.0.

  \begin{table}[t]
    \centering
    \small
    \setlength{\tabcolsep}{4pt}
    \begin{tabular*}{\columnwidth}{@{\extracolsep{\fill}}lrrrr@{}}
      \toprule
      \textbf{ST Model} & \textbf{CER} $\downarrow$ & \textbf{RTF} $\downarrow$ & \textbf{Params (B)} & \textbf{VRAM (GB)} \\
      \midrule
      Tiny & 32.56 & 0.447 & 1.383 & 5.35 \\
      Base & 31.90 & 0.466 & 1.418 & 5.53 \\
      Small & 26.80 & 0.489 & 1.586 & 6.13 \\
      Medium & 23.48 & 0.603 & 2.109 & 8.17 \\
      Large-v2 & 22.68 & 0.674 & 2.889 & 11.29 \\
      \bottomrule
    \end{tabular*}
    \caption{
      \textbf{End-to-end deployment cost on YT-THDC.}
      Each pipeline generates one automatic Mandarin semantic anchor with the indicated upstream ST model, encodes it with mBERT, and performs SAMA-ASR decoding.
      MT-based multilingual-anchor expansion is not included.
    }
    \label{tab:end_to_end_inference_cost}
  \end{table}

  Table \ref{tab:end_to_end_inference_cost} measures the complete practical pipeline with automatic semantic anchors.
  Increasing ST capacity consistently improves downstream CER at the cost of greater latency, model size, and GPU memory.
  The Small translator offers a lower-cost operating point at 0.489 RTF, while the Medium translator obtains the 23.48\% CER used in the main practical comparison at 0.603 RTF.
  All evaluated configurations remain faster than real time, allowing the semantic-anchor generator to be selected according to the desired accuracy--efficiency trade-off.
}

\end{document}